\documentclass[11pt]{article}

\usepackage[final]{acl}

\usepackage{times}
\usepackage{latexsym}
\usepackage{subcaption}
\usepackage{bm}
\usepackage[T1]{fontenc}

\usepackage[utf8]{inputenc}
\usepackage{tikz}
\usetikzlibrary{positioning,fit}
\usepackage{graphicx}
\usepackage{booktabs}
\usepackage{microtype}
\usepackage{xurl}
\usepackage{hyperref} 
\usepackage{inconsolata}

\usepackage{graphicx}
\usepackage{amsmath}

\usepackage{listings}
\usepackage{xcolor}
\usepackage{enumitem}
\usepackage{color}
\usepackage{tabularray}
\usepackage{booktabs}
\usepackage{makecell}

\lstdefinelanguage{json}{
  basicstyle=\ttfamily\small,
  numbers=none,
  showstringspaces=false,
  morestring=[b]"
}
\usepackage{caption}

\definecolor{aftercolor}{RGB}{8,143,143}
\usepackage[table]{xcolor}
\definecolor{highlight}{RGB}{235,240,255}

\definecolor{listing-bg}{RGB}{245,247,250}     
\definecolor{listing-frame}{RGB}{225,225,225}  
\definecolor{listing-number}{RGB}{120,130,140} 
\definecolor{listing-text}{RGB}{20,20,20}      
\definecolor{listing-keyword}{RGB}{0,102,153}  
\definecolor{listing-string}{RGB}{8,143,143}   
\definecolor{listing-comment}{RGB}{128,138,150} 

\lstdefinestyle{cli}{
  backgroundcolor=\color{listing-bg},
  frame=single,
  rulecolor=\color{listing-frame},
  framerule=1.2pt,
  basicstyle=\ttfamily\small\color{listing-text},
  keywordstyle=\color{listing-keyword}\bfseries,
  stringstyle=\color{listing-string},
  commentstyle=\color{listing-comment}\itshape,
  numberstyle=\tiny\color{listing-number},
  numbers=left,
  numbersep=6pt,
  xleftmargin=6pt,
  xrightmargin=6pt,
  breaklines=true,
  breakatwhitespace=true,
  showstringspaces=false,
  postbreak=\mbox{\textcolor{listing-number}{$\hookrightarrow$}\space},
  columns=fullflexible
}

\usepackage[table]{xcolor}
\usepackage{booktabs}
\usepackage{makecell}   
\definecolor{assetcolor}{RGB}{8,143,143}

\title{IndustryAssetEQA: A Neurosymbolic Operational Intelligence System for Embodied Question Answering in Industrial Asset Maintenance}

\author{Chathurangi Shyalika \\
  Artificial Intelligence Institute, \\
  University of South Carolina \\
  USA \\
  \texttt{jayakodc@email.sc.edu} \\\And
  Dhaval Patel \\
  Software Innovation Lab \\
  IBM Yorktown \\
  USA \\
  \texttt{pateldha@us.ibm.com} \\\And
  Amit Sheth \\
  Artificial Intelligence Institute,\\University of South Carolina \\
  Indian AI Research Organization \\
  \texttt{amit@sc.edu, amit@iairo.ai}
  }

\begin{document}
\maketitle
\begin{abstract}
Industrial maintenance environments increasingly rely on AI systems to assist operators in understanding asset behavior, diagnosing failures, and evaluating interventions. Although large language models (LLMs) enable fluent natural-language interaction, deployed maintenance assistants routinely produce generic explanations that are weakly grounded in telemetry, omit verifiable provenance, and offer no testable support for counterfactual or action-oriented reasoning that undermine trust in safety-critical settings. We present \textbf{IndustryAssetEQA}, a neurosymbolic operational intelligence system that combines episode-centric telemetry representations with a Failure Mode and Effects Analysis Knowledge Graph (FMEA-KG) to enable Embodied Question Answering (EQA) over industrial assets. We evaluate on four datasets covering four industrial asset types, including rotating machinery, turbofan engines, hydraulic systems, and cyber–physical production systems. Compared to LLM-only baselines, IndustryAssetEQA improves structural validity by up to $+0.51$, counterfactual accuracy by up to $+0.47$, and explanation entailment by $+0.64$, while reducing severe expert-rated overclaims from 28\% to 2\% ($\approx 93\%$). Code, datasets, and the FMEA-KG are available at: \url{https://github.com/IBM/AssetOpsBench/tree/IndustryAssetEQA/IndustryAssetEQA}.
\end{abstract}

\section{Introduction}

Industrial asset maintenance is rapidly shifting from periodic inspections and offline analysis toward continuous, data-driven decision making \cite{patel2025assetopsbench, seneviratne2018smart}. Modern plants generate large volumes of multivariate telemetry, alert streams, and maintenance records. Operators increasingly rely on AI-based operational intelligence systems to interpret this information~\cite{peres2020industrial, lee2020industrial}. 

Many deployed and prototype systems today use large language models (LLMs) with natural-language interfaces layered over documents, dashboards, and structured data. While these systems produce fluent explanations, practitioner feedback and recent analyses report recurring reliability failures \cite{ji2023survey, schmidtova2025do}: (i) generic explanations that cite textbook failure patterns without grounding in the episode’s sensors or context; \cite{ji2023survey, schmidtova2025do} (ii) missing verifiable provenance as answers rarely cite the supporting time window, sensors, events, or maintenance records, hindering auditability; \cite{Chen2021ImprovingFI} and (iii) non-testable counterfactuals and action suggestions that state qualitative effects without an explicit model of how failure risk changes under interventions. 
These failures show that LLM-centered QA frames maintenance as language generation rather than a sensor-and-risk decision problem; we therefore require time-situated, provenance-backed, risk-anchored QA systems.

The field of \emph{Embodied AI} provides a natural conceptual foundation for addressing this gap. Embodied AI concerns agents with sensors and actuators that perceive the environment, reason about causes, predict the effects of interventions, and act accordingly \cite{fung2025embodied}. This \emph{perception–prediction–reasoning–decision} loop is structurally isomorphic to industrial maintenance workflows: sensing telemetry, diagnosing faults, planning interventions, and executing maintenance actions. 
We argue that industrial maintenance QA should be formulated as an \emph{embodied decision problem} rather than purely textual question answering. In this setting, a system must satisfy four requirements:

\begin{figure*}[!ht]
\centering
\includegraphics[width=1.0\linewidth]{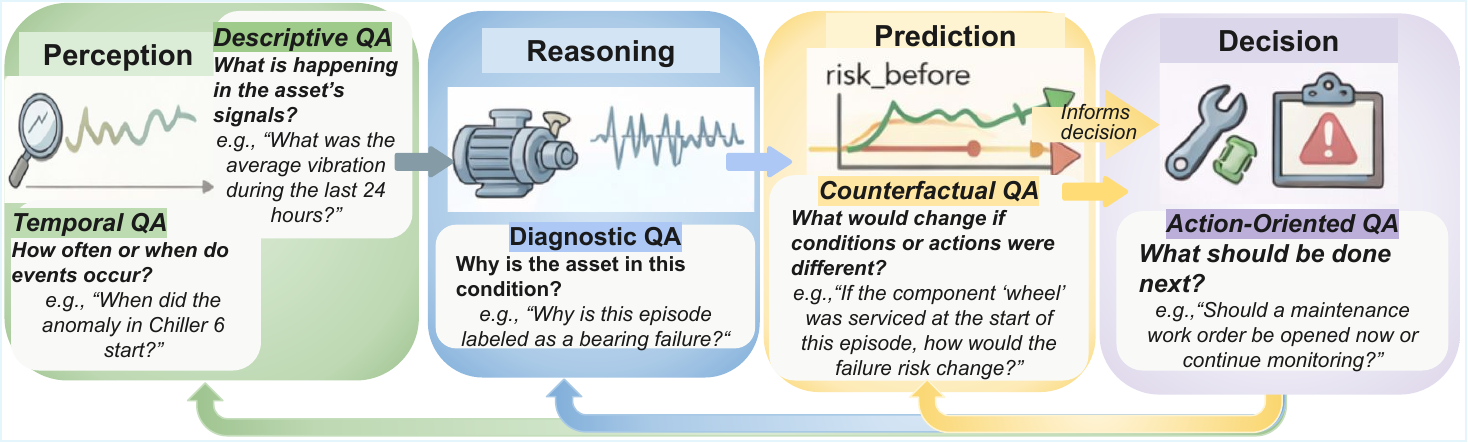}
\caption{QA taxonomy mapped to the perception–reasoning–prediction–decision loop. Forward arrows indicate cognitive dependencies; backward arrows show the embodied interaction loop (decisions → actions → system → observations).}

  \label{fig:qa_types}
\end{figure*}

\begin{itemize}[noitemsep, topsep=0pt, parsep=0pt, partopsep=0pt,
                leftmargin=1.5mm, labelsep=1mm]
    \item \textbf{Time-situated:} Answers must explicitly reference the relevant telemetry window and asset context.
    \item \textbf{Evidence-grounded:} Claims must include verifiable provenance linking to sensors, engineered features, events, and maintenance records.
    \item \textbf{Risk-constrained:} Counterfactual predictions and action recommendations must be derived from an explicit probabilistic risk model.
    \item \textbf{Knowledge-grounded:} Explanations must align with domain semantics and respect structured failure constraints.
\end{itemize}

To operationalize this formulation, we present \textbf{IndustryAssetEQA}, a neurosymbolic operational intelligence system for \emph{Embodied Question Answering} over industrial asset operations. Neurosymbolic integration combines neural learning with symbolic reasoning to enable transparent, knowledge-aligned decision support \cite{sheth2023neurosymbolic}.

Our key contributions are:

\begin{itemize}[noitemsep, topsep=0pt, parsep=0pt, partopsep=0pt,
                leftmargin=1.5mm, labelsep=1mm]
    \item We introduce IndustryAssetEQA, a deployment-oriented embodied QA framework for asset maintenance, and a dataset of structured, time-situated episode facts spanning five QA task types.
    \item We construct a domain-specific Failure Mode and Effects Analysis Knowledge Graph (FMEA-KG) to enable neurosymbolic grounding of industrial failure semantics across assets and components.

    \item We implement a parametric intervention-style risk estimator trained on time-series telemetry, which yields testable estimates of risk direction and magnitude under maintenance interventions.

    \item We design provenance- and structure-aware evaluation protocols that measure structural validity, evidence grounding, temporal alignment, and counterfactual consistency to assess deployable reliability in industrial QA.
\end{itemize}

\section{Related Work}
\textbf{Industrial AI and QA for Asset Maintenance.}
Prior work spans autonomous industrial agents and evaluation \cite{patel2025assetopsbench}, domain-specific QA \cite{Terziyan2025ACD}, and decision-support systems based on knowledge-enhanced LLMs for manufacturing \cite{Chatterjee2022AutomatedQF,Lee2023AQM,Lin2025FineTunedTL,Constantinides2025TowardsBG}. These systems typically operate on static documents and global datasets.

\textbf{Time Series and Sensor-Grounded Question Answering.}
Recent work reformulates temporal reasoning as time series QA, enabling querying over numerical sequences, multimodal signals, and temporal events \cite{constantinidesfailuresensoriq,Su2025ChainofthoughtRA,Kong2025TimeMQATS,Chen2025MTBenchAM,Wang2025ITFormerBT,Uddin2024UnSeenTimeQATQ}. These approaches introduce benchmarks and models for structured reasoning, multimodal fusion, and error correction, including reviewer-based CoT verification \cite{Su2025ChainofthoughtRA}, large-scale multitask time series datasets \cite{Kong2025TimeMQATS,Wang2025ITFormerBT}, and cross-modal temporal QA combining text and time series \cite{Chen2025MTBenchAM}. However, even strong LLMs struggle with long-range dependencies, causal interpretation, parallel event reasoning, and zero-shot temporal generalization \cite{Uddin2024UnSeenTimeQATQ,Merrill2024LanguageMS}. 

\begin{figure*}[!ht]
\centering
\includegraphics
[width=0.85\linewidth]
{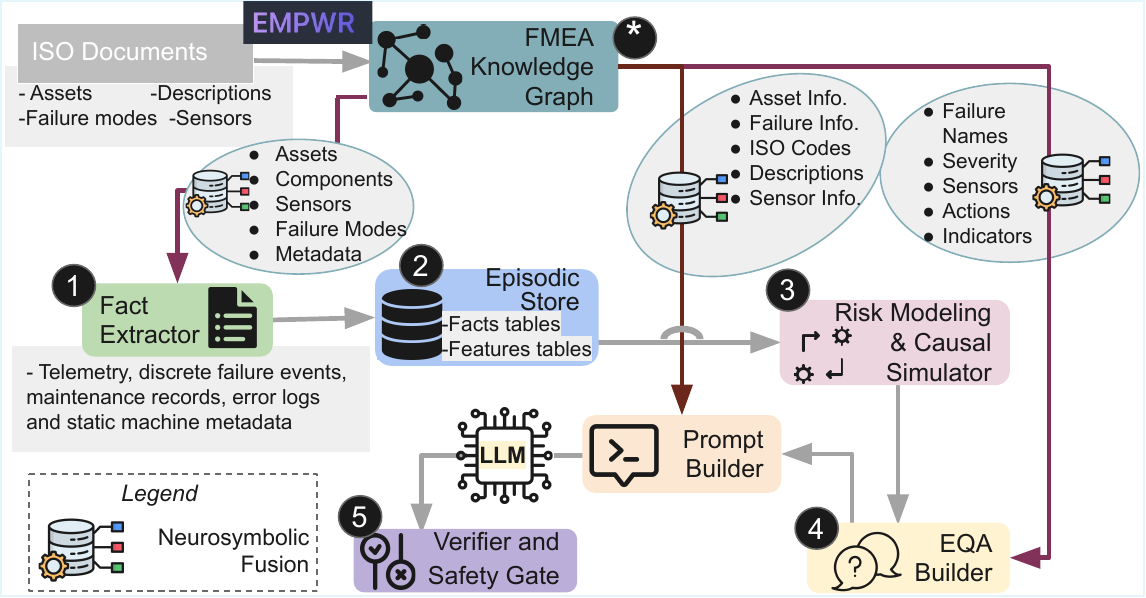}
    \caption{IndustryAssetEQA Architecture integrating Fact Extractor, Episodic Store, FMEA-KG, Causal Simulator, Verifier and Safety Gate}
    \label{fig:archi_a}
\end{figure*}

\textbf{Embodied Question Answering and Decision-Centric AI.} Prior work on embodied question answering extends perception-driven QA to interactive 3D environments \cite{Li2025EmbodiedIF}, incorporating external knowledge \cite{Tan2021KnowledgeBasedEQ}, planning \cite{Ginting2025EnterTM}, and domain-specific benchmarks for industrial and safety-critical settings \cite{Li2025EmbodiedIF,Tan2021KnowledgeBasedEQ,Ginting2025EnterTM,Majumdar2024OpenEQAEQ,Li2025IndustryEQAPT,Li2025IndustryNavES,Zhang2025EmbodiedII}. While these approaches advance embodied perception, memory, and language grounding, they largely evaluate semantic, spatial, or temporal understanding from observations or simulators, and do not explicitly verify the correctness or risk consistency of operational decisions.

\section{Industrial Setting and QA Task Taxonomy}
\label{sec:problem_setting}
We target industrial asset maintenance (manufacturing, utilities, process plants) where operators inspect multivariate telemetry, alerts, and maintenance logs to make time-sensitive decisions. Each query is framed as \textit{episode-centric (an asset + a concrete time window)} and evaluates whether answers are (i) semantically grounded in episodic and KG evidences, and (ii) when relevant consistent with a simple, data-driven risk model for interventions. The framework covers five compact QA types and are directly aligned with perception $\rightarrow$ reasoning $\rightarrow$ prediction $\rightarrow$ decision pipeline as visualized in Figure \ref{fig:qa_types}.

\begin{figure}[!ht]
\centering
\includegraphics
[width=0.75\linewidth]{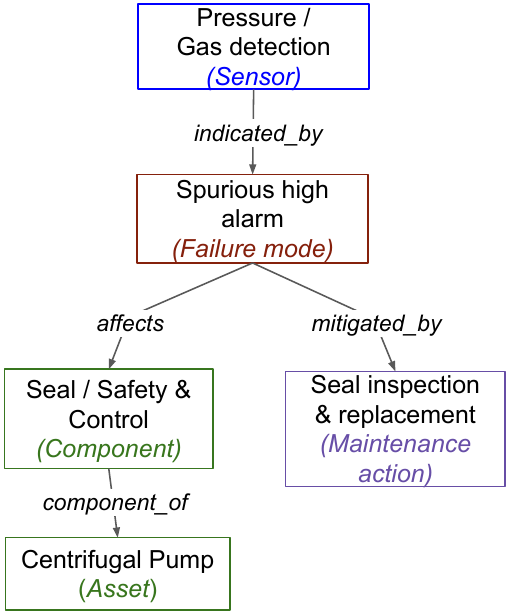}
 \caption{Representative FMEA-KG fragment}
    \label{fig:archi_b}
\end{figure}

\section{IndustryAssetEQA System Overview}
\label{sec:overview}

Figure~\ref{fig:archi_a} illustrates the overall architecture that includes the core components of IndustryAssetEQA. 

\paragraph{FMEA Knowledge Graph and Neurosymbolic Fusion.} We integrate domain knowledge derived from expert-curated Failure Mode and Effects Analysis (FMEA) specifications from ISO-documentation in the form of a machine-interpretable knowledge graph using the EMPWR platform \cite{Yip2024TheEP}. The graph comprises \textbf{63 failure modes} mapped to \textbf{9 asset categories} and their component types (Appendix \ref{sec:fmeakg}). Nodes represent asset classes, subcomponents, failure modes, sensor abstractions, and maintenance actions, and edges capture relations such as \emph{affects}, \emph{component\_of}, \emph{indicated\_by}, and \emph{mitigated\_by} (Figure \ref{fig:archi_b}). Failure mode nodes are associated with expected telemetry signatures, severity and occurrence metadata, and admissible interventions. FMEA-KG serves as a symbolic grounding (via neurosymbolic fusion) layer that interfaces with neural components at well-defined points in the pipeline using \emph{rdflib} \cite{rdflib}. 
Extracted nodes and relations are used as semantic context, so LLM outputs are auditable, domain-valid explanations that link sensor patterns to legitimate failure modes and actions.


\paragraph{Fact Extractor.}
It converts heterogeneous time series data into structured episode-level facts. For each failure occurring at time \(t_f\) on machine \(m\), the extractor defines a configurable historical window \([t_f-\Delta,\,t_f]\) and aggregates all telemetry samples in that interval. For every numeric sensor signal \(s\) it computes a compact set of summary descriptors \(\mathcal{F}_s=\{\mu_s,\sigma_s,\min_s,\max_s,\mathrm{trend}_s\}\), where \(\mu_s\) and \(\sigma_s\) are the sample mean and standard deviation, and \(\mathrm{trend}_s\) is the slope of a least-squares linear fit to the samples in the window. It forms contextual features such as error event counts, number of distinct error types, and component-specific \(\emph{hours\_since\_last\_maintenance}\). Static attributes (e.g., machine age and one-hot model indicators) are appended to the feature vector. Healthy episodes are sampled by selecting center timestamps \(t\) with no failures in a subsequent horizon \([t,\,t+H]\) and building the corresponding history window \([t-\Delta,\,t]\). Each episode is labeled (``failure'' or ``healthy''), enriched with FMEA-KG (asset profile, failure profile, sensor semantics), and emitted as a JSONL record with full provenance that include source files, time ranges, and counts (Example in Appendix \ref{sec:fact_extraction}).



\paragraph{Episodic Store.}
This is a lightweight persistence and retrieval layer for episode-level facts. Serialized facts are ingested and stored in a SQLite database with a two-table schema: a \texttt{facts table} that retains each episode's full serialized metadata and a \texttt{features table} that decomposes each episode's feature vector into atomic \((\emph{feature\_name},\ \emph{feature\_value})\) rows. Each episode \(f\) is indexed by a stable identifier \(\text{fact\_id}\) and associated metadata (e.g., \(\text{asset\_id}\), \(\text{label}\), source file, and row index). Storing features in an indexed, per-feature form enables efficient numeric queries and threshold searches of the form \emph{\(\{\, f_i \mid x_{i,j} \ \bowtie\ \tau \,\}\)},
where \(x_{i,j}\) denotes the value of feature \(j\) in episode \(i\), \(\tau\) is a numeric threshold, and \(\bowtie\in\{<,\le,=,\ge,>\}\). The store exposes a compact API for retrieval by asset, label, feature constraints, and balanced sampling across labels (See Appendix \ref{sec:episodic_store}). 



\paragraph{Risk Modeling and Causal Simulator.}
Implements a compact, data-driven risk model and a parametric counterfactual estimator. Given an episode fact represented by a fixed-length feature vector $\bm{x}$, the trainer fits a multinomial logistic regression to approximate the conditional distribution $P(y\mid\bm{x})$ over discrete episode labels $y$ (e.g., \texttt{failure modes} and \texttt{healthy}). At inference the simulator interprets a maintenance or sensor intervention as an explicit substitution $\bm{x}\mapsto\bm{x}^{\mathrm{do}}$ (i.e., setting one or more coordinates to new values) and computes the pre- and post-intervention risks as:
\begin{equation}\label{eq:r_before_after}
\begin{aligned}
r_{\mathrm{before}}
&= 1 - P\bigl(y=\mathrm{healthy}\mid \bm{x}\bigr),\\[4pt]
r_{\mathrm{after}}
&= 1 - P\bigl(y=\mathrm{healthy}\mid \bm{x}^{\mathrm{do}}\bigr),
\end{aligned}
\end{equation}
so that the estimated (signed) risk change is:
\begin{equation}\label{eq:delta_r}
\Delta r \;=\; r_{\mathrm{after}} - r_{\mathrm{before}}.
\end{equation}
The simulator returns class-probability maps and a lightweight confidence heuristic based on probability extremity (Details in Appendix~\ref{sec:causal_sim}). It outputs the pre- and post-intervention probability vectors
\[
\bigl\{P(y=k \mid \bm{x})\bigr\}_{k}
\qquad\text{and}\qquad
\bigl\{P(y=k \mid \bm{x}^{\mathrm{do}})\bigr\}_{k}.
\]

\paragraph{EQA Builder.} 
Given an episodic fact, EQA Builder deterministically constructs task-specific questions, derives answers from telemetry-derived features and the risk model, and assembles structured reasoning traces with provenance. Where available, it enriches QA instances by querying the FMEA-KG to retrieve failure-mode metadata (e.g., severity, associated sensors, typical indicators, recommended actions) and asset descriptors (equipment category and class)(Samples in Appendix \ref{app:examples}). 


\paragraph{Prompt Builder.}\label{sec:prompt-builder}
A deterministic, safety-aware interface converting an episodic fact and QA into scoped LLM prompts. For each fact, it renders a compact, human- and machine-readable evidence block containing \texttt{provenance} (\texttt{fact\_id, asset, source, row}), \texttt{the episode window}, \texttt{key diagnostic features}, and \texttt{FMEA context} when available. The builder then maps the QA task type to a short, task-specific instruction and produces an explicit user instruction that (a) asks the question, (b) constrains the LLM to use only the supplied evidence, (c) requires a strict, machine-parsable JSON output with keys \texttt{direct\_answer}, \texttt{reasoning\_answer}, \texttt{provenance}, and \texttt{confidence}, and (d) for counterfactual and action tasks mandates a numeric \texttt{counterfactual} object containing \texttt{risk\_before}, \texttt{risk\_after} and a direction label consistent with those numbers (Example prompt in Appendix \ref{sec:prompt_builder}). 

\paragraph{Verifier and Safety Gate.}

A deterministic assessment module that validates the output, and cited metadata against the Episodic Store and returns a structured diagnostic bundle of issues. The safety gate is a downstream policy layer that consumes verifier’s report and enforces admissibility. It flags answers that fail checks, logs incidents for audit, and routes problematic cases to human review.


\paragraph{Runtime Deployment Flow.}
In a deployed setting, IndustryAssetEQA operates as a closed-loop operational intelligence service. An operator submits a natural-language query through a maintenance interface (e.g., a dashboard or a Computerized Maintenance Management System (CMMS) frontend), which is first classified by question type. The system then retrieves relevant episodic facts and FMEA knowledge, invokes the risk simulator when counterfactual reasoning is required, and generates a structured, provenance-enforced answer. Before any recommendation is surfaced, a safety and action gate validates evidence consistency and risk thresholds. Approved recommendations (e.g., work-order creation or escalation) are then forwarded to downstream maintenance systems, while low-confidence cases are flagged for human review.

\section{Datasets and QA Task Design}
\label{sec:data}
\begin{table}[!ht]
\centering
\footnotesize
\setlength{\tabcolsep}{1pt} 
\renewcommand{\arraystretch}{0.92} 

\begin{tabular}{@{}l@{\hskip -3pt}rrrrrr@{}}
\toprule
\makecell{\textbf{Dataset} \\ \textbf{(\& Asset)}} & \textbf{Episodes} & \textbf{Desc.} & \textbf{Temp.} & \textbf{Diag.} & \textbf{CF.} & \textbf{Act.} \\
\midrule
\makecell[l]{Microsoft PdM \\ {\color{assetcolor}(Rotating machinery)}} &
  5716 & 5716 & 5716 & 5716  & 761  & 902 \\ \hline
\makecell[l]{C-MAPSS \\ {\color{assetcolor}(Turbofan engines)}} &
  4842 & 4842 & 4842 & 4842 & -  & - \\ \hline
\makecell[l]{Genesis CPS \\ {\color{assetcolor}(Cyber--physical}\\ \color{assetcolor}systems)} &
 478 & 120 & 478 & 214 & 210 & 23  \\ \hline
\makecell[l]{Hydraulic systems \\ {\color{assetcolor}(Hydraulic test rig)}} &
 2205  & 2205  & 2205  & 50  & 2184  & 2205  \\
\midrule
\textbf{Total} & 13241 & 12883 & 13241 & 10822 & 3155 & 3130 \\
\bottomrule
\end{tabular}
\caption{Datasets, asset statistics and number of QA instances per task category. ``Desc.'' = descriptive, ``Temp.'' = temporal, ``Diag.'' = diagnostic, ``CF.'' = counterfactual, ``Act.'' = action-oriented.}
\label{tab:task_stats}
\end{table}

We use four industrial datasets spanning predictive maintenance \cite{azure_pdm_kaggle}, degradation modeling \cite{saxena2008damage, nasa_cmapss_opendata}, cyber--physical systems \cite{von2018anomaly} and hydraulic systems \cite{hydraulic_zenodo_1323611, helwig2015condition, helwig2015d8}. Table \ref{tab:task_stats} includes a summary of them, and the episodes and QAs were validated by SMEs and used as the ground truth for the experiments. Detailed dataset descriptions provided in Appendix~\ref{sec:datasets}. These ground-truth represent expert-approved interpretations of the provided evidence under a closed-world assumption.







\section{Evaluation Methodology}

\paragraph{Evaluation Overview.}

IndustryAssetEQA is evaluated as an operational QA system rather than as a purely linguistic model. Each evaluation instance consists of a single QA query grounded in a specific episodic fact. For each QA instance, we generate one model response under each system configuration (\textit{baselines}: LLM-only, episodic evidence, KG-grounding, provenance-enforced; and the \textit{full IndustryAssetEQA}) (Details in Appendix \ref{sec:baselines}). We evaluate using black-box API access to \texttt{GPT-4o-mini} and \texttt{Claude Sonnet 4}. 
Models are prompted with a structured episode-level fact, optional FMEA-KG context, and a strict JSON output contract, without access to ground-truth data.

\paragraph{Metrics.}We evaluate IndustryAssetEQA for \emph{reliability} and \emph{operational validity} in industrial settings, rather than surface-level answer fluency. In safety- and cost-critical scenarios, a QA system must produce structurally valid outputs (\emph{Struct.OK}), ground its answers in verifiable episodic evidence (\emph{Prov.OK}), and generate counterfactual predictions consistent with an explicit risk model (\emph{CF Acc.}). Beyond final-answer correctness, we assess the \emph{semantic faithfulness} of generated explanations. Direct answers are evaluated against SME-validated references using task-appropriate accuracy measures (\emph{Label Cons.}), while descriptive and temporal questions are evaluated via exact and tolerance-based matching. Reasoning quality is evaluated independently of gold explanations by checking entailment to episodic and symbolic evidence (\emph{Entail.Pass}) and verifying extracted claims against the episodic store and FMEA knowledge graph (\emph{Claim Prec.}). Expert evaluation is used for action-oriented questions where no single oracle exists. Metrics are computed per QA instance and then aggregated across all datasets and applicable task types (See Appendix \ref{sec:metrics}).

\section{Experiments and Results}

\begin{table*}[!ht]
\centering
\small
\setlength{\tabcolsep}{4pt}
\resizebox{\textwidth}{!}{%
\begin{tabular}{l*{6}{c}*{6}{c}}
\toprule
\textbf{Method} &
\multicolumn{6}{c}{\textbf{\texttt{GPT-4o-mini}}} & \multicolumn{6}{c}{\textbf{\texttt{Claude Sonnet 4}}} \\
\cmidrule(lr){2-7}\cmidrule(lr){8-13}
 & \shortstack{\textbf{Struct.}\\\textbf{OK} $\uparrow$} &
   \shortstack{\textbf{Prov.}\\\textbf{OK} $\uparrow$} &
   \shortstack{\textbf{Label}\\\textbf{Cons.} $\uparrow$} &
   \shortstack{\textbf{CF}\\\textbf{Acc.} $\uparrow$} &
   \shortstack{\textbf{Entail.}\\\textbf{Pass} $\uparrow$} &
   \shortstack{\textbf{Claim}\\\textbf{Prec.} $\uparrow$} &
   \shortstack{\textbf{Struct.}\\\textbf{OK} $\uparrow$} &
   \shortstack{\textbf{Prov.}\\\textbf{OK} $\uparrow$} &
   \shortstack{\textbf{Label}\\\textbf{Cons.} $\uparrow$} &
   \shortstack{\textbf{CF}\\\textbf{Acc.} $\uparrow$} &
   \shortstack{\textbf{Entail.}\\\textbf{Pass} $\uparrow$} &
   \shortstack{\textbf{Claim}\\\textbf{Prec.} $\uparrow$} \\
\midrule
LLM-only QA
 & 0.42 & 0.47 & 0.62 & 0.45 & 0.08 & 0.12
& 0.39 & 0.44 & 0.59 & 0.44 & 0.1 & 0.12 \\
LLM + Episodic
 & 0.52 & 0.62 & 0.71 & 0.45 & 0.23 & 0.25
& 0.54 & 0.65 & 0.73 & 0.44 & 0.27 & 0.28 \\
LLM + Episodic + KG
 & 0.52 & 0.62 & 0.73 & 0.45 & 0.56 & 0.51
  & 0.55 & 0.62 & 0.73 & 0.44 & 0.58 & 0.54 \\
+ Provenance-Enforced
 & 0.82 & 0.83 & 0.89 & 0.45 & 0.63 & 0.59
& 0.80 & 0.84 & 0.87 & 0.44 & 0.61 & 0.60 \\
\rowcolor{highlight}
\textbf{Full IndustryAssetEQA}
 & \textbf{0.88} & \textbf{0.89} & \textbf{0.94} & \textbf{0.88} & \textbf{0.72} & \textbf{0.67}
 & \textbf{0.90}   & \textbf{0.89}   & \textbf{0.95}   & \textbf{0.91}   & \textbf{0.78}   & \textbf{0.74} \\

\bottomrule
\end{tabular}%
} 
\caption{
Per-model results for answer correctness, grounding, and explanation faithfulness. \emph{Struct.OK} = structural validity, \emph{Prov.OK} = evidence grounding, \emph{Label Cons.}= KG-normalized agreement between the predicted diagnostic label and the SME-validated reference, \emph{CF Acc.}= agreement with the expert-reviewed surrogate model on counterfactual direction, \emph{Entail. Pass} = NLI (\texttt{FacebookAI/roberta-large-mnli}) $\geq$ 0.80; and \emph{Claim Prec.} = measure explanation faithfulness against episodic evidence and the KG.
}
\label{tab:answer_and_reasoning_metrics_by_model}
\end{table*}
We evaluate IndustryAssetEQA through a set of focused research questions (RQs). 

\paragraph{RQ1: Does Embodied Question Answering Improve Answer Reliability?} We measure reliability via structural validity, provenance accuracy, diagnostic label consistency, counterfactual direction accuracy, and explanation faithfulness(Table~\ref{tab:answer_and_reasoning_metrics_by_model}). Both models follow the same qualitative trajectory. LLM-only baseline performs poorly. Adding episodic evidence improves grounding and explanations, while leaving counterfactual accuracy unchanged, and neurosymbolic fusion with the FMEA-KG raises explanation faithfulness and label consistency. Provenance enforcement produces the largest jump in deployable reliability, and the full IndustryAssetEQA yields the best results (GPT: Struct.OK: 0.88, Prov.OK: 0.89, Label Cons.: 0.94, CF Acc.: 0.88; Claude: Struct.OK: 0.90, Prov.OK: 0.89, Label Cons.: 0.95, CF Acc.: 0.91). Episodic grounding and KG context improve semantic faithfulness, whereas provenance enforcement and simulator integration drive the principal gains required for deployable reliability.

\paragraph{Statistical Significance Analysis.}
We assess whether differences between model variants are statistically significant using McNemar’s test \cite{laerd_mcnemar_spss, mcpanalytics_mcnemar_guide} for predictions of the Microsoft PdM dataset (Table \ref{tab:significance}). For descriptive, diagnostic, and counterfactual QA, the differences are statistically significant (p $<$ 0.05), while for temporal and action-oriented QA, no statistical significance observed (p $>$ 0.1). For action-oriented QA, which is central to decision-making, the two variants exhibit nearly identical behavior (p = 0.93). This indicates that, model choice alone is insufficient to address challenges in action-oriented reasoning, underscoring the need for structured, provenance-aware components.

\begin{table}[!ht]
\centering
\small
\setlength{\tabcolsep}{6pt}
\begin{tabular}{lccc}
\toprule
\textbf{Question type} & \textbf{\# Questions} & \textbf{p-value} & \textbf{Significance} \\
\midrule
Descriptive      & 5716 & 0.003 & Yes \\
Diagnostic       & 5716 & 0.021 & Yes \\
Temporal         & 5716 & 0.14  & No  \\
Counterfactual   & 761  & 0.04  & Yes \\
Action-oriented  & 902  & 0.93  & No  \\
\bottomrule
\end{tabular}
\caption{Statistical significance analysis using McNemar’s test across QA types on the Microsoft PdM dataset.}
\label{tab:significance}
\end{table}

\paragraph{RQ2: Can Models Predict the Effects of Actions on Physical Asset Risk?} We evaluate counterfactual reasoning using direction accuracy, which measures whether system outputs are directionally consistent with expert-reviewed surrogate risk model under hypothetical maintenance interventions(Table~\ref{tab:answer_and_reasoning_metrics_by_model}). In an embodied QA framing, this tests whether answers are consistent with modeled action–effect relationships for the asset rather than merely producing plausible-sounding text. Baselines without simulator access perform near chance ($\approx 0.45$). IndustryAssetEQA achieves a counterfactual direction accuracy of $0.88$ (GPT) and $0.91$ (Claude), indicating strong agreement with expert-reviewed surrogate model. These results suggest that providing an explicit surrogate intervention model materially improves directional consistency between proposed actions and predicted effects.

\paragraph{RQ3: Which Architectural Components Enable Embodied, Decision-Grounded QA?}

\begin{table}[!ht]
\centering
\small
\setlength{\tabcolsep}{1.5pt}
\begin{tabular}{lccc}
\toprule
\textbf{Configuration} & \textbf{Entail.Pass $\uparrow$} & \textbf{Full Pass $\uparrow$} & \textbf{CF Acc. $\uparrow$} \\
\midrule
\rowcolor{highlight}
Full IndustryAssetEQA & \textbf{0.72} & \textbf{0.89} & \textbf{0.88} \\
w/o Risk simulator & 0.59 & 0.72 & 0.49 \\
w/o Provenance \\enforcement & 0.42 & 0.19 & 0.81 \\
w/o FMEA-KG  & 0.59 & 0.35 & 0.61 \\
w/o Episodic memory & 0.27 & 0.36 & 0.34 \\
\bottomrule
\end{tabular}

\caption{Ablation results for \texttt{GPT-4o-mini}. Full pass= Avg(Struct.OK+Prov.OK+Label Cons.)}

\label{tab:ablation}

\end{table}

We measure each component's contributions via controlled ablations (Table~\ref{tab:ablation}, Appendix \ref{sec:ablation}).
Removing the risk simulator primarily breaks counterfactual reasoning (CF Acc. 0.88 → 0.49) while only moderately lowering entailment and full-pass rates, showing simulation is essential for reliable what-if judgments. Disabling provenance enforcement causes the largest collapse in deployable reliability (Full Pass 0.89 → 0.19) and strongly reduces explanation faithfulness (Entail.Pass 0.72 → 0.42), indicating that numeric predictions without verifiable grounding are unsafe. Removing episodic memory degrades all metrics (Entail.Pass 0.72 → 0.27; Full Pass 0.89 → 0.36; CF Acc. 0.88 → 0.34), and dropping FMEA-KG lowers semantic alignment and reliability (Entail.Pass 0.72 → 0.59; Full Pass 0.89 → 0.35), so episodic grounding, provenance, KG context, and simulation are complementary and jointly required for deployable embodied QA.

\paragraph{RQ4: How Reliable and Semantically Valid is the FMEA-KG Used for Grounding?}
We assess the quality of the FMEA-KG along three axes: 
\textit{(i) expert validation}:  the schema and derived QA instances are reviewed by SMEs (Section \ref{sec:data}); 
\textit{(ii) functional validation}: ablations show consistent performance degradation without the KG (e.g., lower entailment and full-pass rates in Tables \ref{tab:answer_and_reasoning_metrics_by_model} and \ref{tab:ablation}); and 
\textit{(iii) consistency checks}: KG-derived outputs are verified against episodic telemetry via provenance constraints (Section \ref{sec:overview} and Table \ref{tab:answer_and_reasoning_metrics_by_model}).

To further quantify quality, we perform triple-level verification using text entailment against ISO-style specifications. Among 1004 candidate triples, $96.0\%$ are judged valid. Support is strongest for semantic relations such as \emph{description} ($\approx 76.7\%$) and \emph{involves} ($\approx 71.4\%$), while structurally weaker relations (e.g., sample/example links, $<10\%$) exhibit higher noise, indicating areas for refinement. In a 50-triple spot check, $92\%$ of triples achieve high entailment confidence ($>0.6$), providing additional evidence of semantic consistency, although some cases remain near the decision boundary. These results indicate that the FMEA-KG reliably captures domain-relevant semantics required for explanation grounding and reasoning, while highlighting limitations in structurally weaker relations that do not directly impact downstream QA performance.



\paragraph{RQ5: Do Expert Evaluations Reveal Failure Modes Not Captured by Automatic Metrics?}

\begin{figure}[!ht]
\centering

\includegraphics[width=\linewidth]{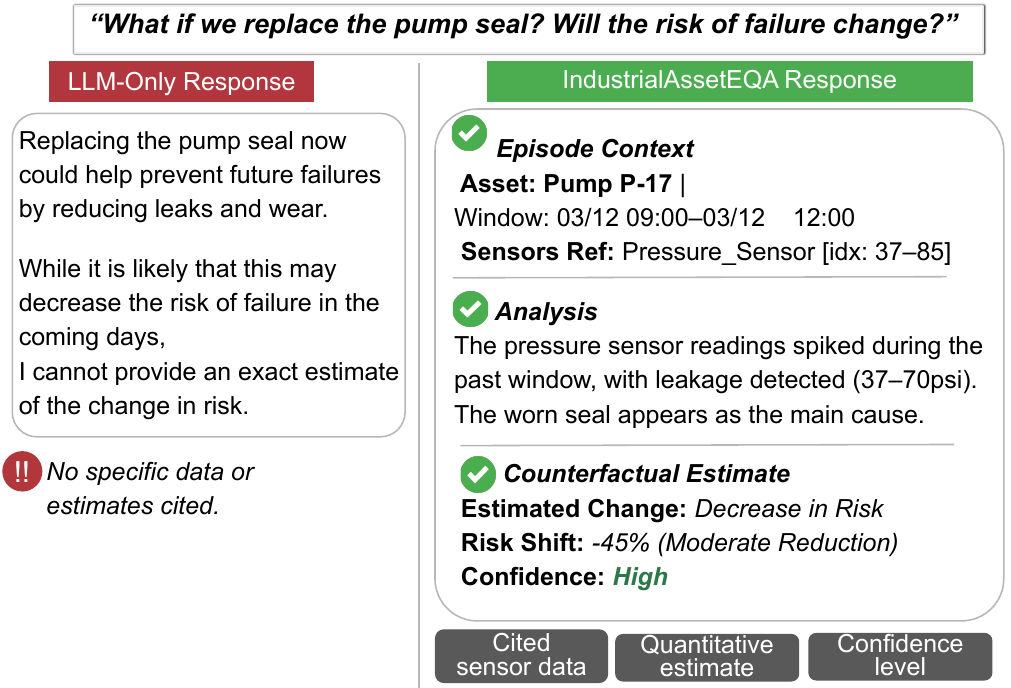}

\caption{Qualitative comparison on a diagnostic and counterfactual query. 
}
\label{fig:qualitative_example}
\end{figure}

We conducted a blinded expert evaluation\footnote{Survey instrument (sample): \url{https://github.com/IBM/AssetOpsBench/blob/IndustryAssetEQA/IndustryAssetEQA/UserStudy.pdf}.} over 22 QA pairs assessed by 10 experts. QA pairs were stratified across four datasets and five task types, with uniform allocation per category. Judges assessed each QA pair on answerability, data grounding (1--5), ambiguity (1--5), scope correctness, and overclaim severity. IndustryAssetEQA outputs were judged answerable in 97\% of cases, compared to 46\% for the LLM-only baseline (absolute gain = 45\%). Mean data grounding scores were significantly higher for IndustryAssetEQA (4.5 $\pm$ 0.6) than for LLM-only outputs (3.0 $\pm$ 0.9; paired $t$-test, $p < 0.001$). Severe overclaims occurred in 2\% of responses, compared to 28\% for LLM-only. Inter-annotator agreement was substantial (Fleiss’ $\kappa = 0.63$). These expert judgments corroborate the automatic metrics and demonstrate that provenance enforcement, task-scoped prompting, and simulator alignment materially reduce unsafe and unsupported claims in industrial QA. Figure~\ref{fig:qualitative_example} provides a representative qualitative comparison between LLM-only QA and IndustryAssetEQA on a diagnostic and counterfactual query. The LLM-only response offers a plausible but unsupported explanation without quantitative risk estimates, whereas IndustryAssetEQA cites episode-level sensor data, provides an explicit counterfactual risk shift, and reports a calibrated confidence level.

\section{Conclusion}


IndustryAssetEQA reframes industrial QA as an embodied decision problem and is engineered for live deployment. We are integrating it as a reliability layer into enterprise agents (e.g., SAP Joule Agents), deploying a controlled pilot to measure latency, operator agreement, escalation behavior, reduction in unsupported claims, and extending the integration for multi-asset production deployment.

\clearpage

\section*{Limitations}
We have identified several limitations of our work. 
\begin{itemize}[noitemsep, topsep=0pt, parsep=0pt, partopsep=0pt,leftmargin=1.5mm, labelsep=1mm]
    \item \textbf{Counterfactual modeling:}
    The counterfactual module is a lightweight parametric intervention estimator rather than a fully identified structural causal model. Its outputs are surrogate risk estimates, not proven causal effects.

\item \textbf{FMEA-KG quality and coverage:}
The KG captures core failure semantics but exhibits variability across relation types, with structurally weaker links being noisier. It is domain-level and may not capture asset-specific variations and rare failure patterns.

    \item \textbf{Fixed episode windowing:}
    Episodes use fixed windows and may miss long-horizon or multi-scale precursors. Adaptive windowing is left to future work.
    
    \item \textbf{Evaluation scope and pilot plan:}
    Evaluation so far is offline on four benchmark datasets. We are preparing a controlled pilot integration to measure operational impact (latency, operator agreement, escalation rate, and incidence of unsupported claims) and to collect data to empirically calibrate counterfactual estimates.
    
    \item \textbf{System overhead and hosting:}
    The full pipeline adds engineering and runtime overhead relative to LLM-only systems. The design supports on-premises, hybrid, or Virtual Private Cloud (VPC) deployment and configurable operating points (confidence thresholds, retraining cadence) to manage cost, latency, and privacy trade-offs.
\end{itemize}

\section*{Ethical Considerations}
This work targets decision support in safety- and cost-critical industrial maintenance settings, where incorrect or overconfident recommendations could lead to operational risk. To mitigate this, IndustryAssetEQA is designed as an advisory system; it enforces explicit evidence citation, provenance verification, and simulator-grounded counterfactuals, and routes low-confidence or inconsistent outputs to human review rather than automating actions. The system does not learn from personal data, does not involve human subjects beyond expert annotation, and operates on publicly available or industrial benchmark datasets. The deployment requires careful integration with organizational safety policies, clear communication of uncertainty, and human-in-the-loop oversight to prevent misuse or over-reliance on automated recommendations.

\section*{Acknowledgments}
This work was supported in part by NSF grant \#2119654, ``RII Track 2 FEC: Enabling Factory to Factory (F2F) Networking for Future Manufacturing''. It is also part of a collaborative effort enabled by the open-source project AssetOpsBench (IBM).

\bibliography{custom}
\clearpage

\section*{Appendix Overview}
\label{app:overview}

This appendix details the data processing, system components, evaluation, and reproducibility of IndustryAssetEQA. It is organized as follows:

\begin{itemize}[leftmargin=*]

    \item \textbf{Appendix \ref{sec:fmeakg}: FMEA Knowledge Graph Construction.} Describes how the domain-level FMEA knowledge graph is constructed from ISO-style specifications using the EMPWR platform.
 
    \item \textbf{Appendix \ref{sec:fact_extraction}: Fact Extraction from Time Series Telemetry.} Describes the deterministic time series extractor used to convert raw telemetry, error logs, and maintenance records into episode-level facts. Also includes representation of a sample fact generated.
    
    \item \textbf{Appendix \ref{sec:episodic_store}: Episodic Store (Ingest and Query Layer).} Details the SQLite-backed episodic store, its schema, ingestion pipeline, query APIs, and how it supports deterministic retrieval and verification.

    \item \textbf{Appendix \ref{sec:causal_sim}: Risk Modeling and Causal Simulator.} Provides implementation details, assumptions, and limitations of the parametric risk model and counterfactual estimator used for ``what-if'' reasoning, including the confidence heuristic.

    \item \textbf{Appendix \ref{app:examples}: Example QA Instances.} Presents representative generated QA instances across different QA tasks to illustrate the FMEA-KG enrichment and structured answer format.
    
    \item \textbf{Appendix \ref{sec:prompt_builder}: Prompt Builder and Prompt Templates.} Documents the deterministic prompt construction process, including system and user prompt structure, evidence rendering, task-specific instructions, and JSON output contracts.

    \item \textbf{Appendix \ref{sec:datasets}: Datasets Used.} Describes the industrial datasets employed in evaluation (PdM, C-MAPSS, Genesis CPS, and Hydraulic Systems), their characteristics, and how episodic windows and QA instances are derived.
    
    \item \textbf{Appendix \ref{sec:evaluation}: Evaluation.} Specifies the evaluation metrics, baselines, ablation design, and automatic verification procedures.
\end{itemize}

\appendix

\section{Appendix: FMEA Knowledge Graph Construction.}
\label{sec:fmeakg}
The FMEA knowledge graph is constructed from ISO-style failure mode specifications and expert-curated maintenance documentation using the EMPWR platform~\cite{Yip2024TheEP}. EMPWR provides a structured workflow for ingesting semi-structured documents, defining domain ontologies, validating entity and relation consistency, and managing KG evolution over time. We use EMPWR to define a reusable asset-centric schema covering asset classes, components, failure modes, sensors, and maintenance actions, and to populate this schema with FMEA-aligned metadata.

The FMEA-KG comprises \textbf{63 distinct failure modes} (e.g., \texttt{AIR} - Abnormal instrument reading, \texttt{ELP} - External leakage, \texttt{OHE} - Overheating, \ldots, \texttt{STU} - Stuck) mapped to component- and asset-level concepts. The metadata associates each failure mode with multiple equipment classes and components across  \textbf{nine asset categories} (\emph{drilling equipment, marine equipment, electrical systems, mechanical systems, rotating machinery, safety and control, subsea equipment, well completion, and well intervention}). The final graph includes \textbf{210 entities} (e.g., Subsea pipelines (\emph{assets}), Pressure or Temperature detection (\emph{sensors}), Heating failure (\emph{Failure modes})) and \textbf{1004 relationships} (e.g., \emph{affects}, \emph{component\_of}, \emph{indicated\_by}, and \emph{mitigated\_by}). The KG is \emph{domain-level} rather than dataset-specific. It is constructed independently of any particular dataset and reused across all evaluated datasets. Each dataset activates a subset of the KG corresponding to its asset type (e.g., rotating machinery, hydraulic systems, turbofan engines), sensors, and failure modes. This design supports cross-dataset generalization while ensuring that explanations remain semantically aligned with domain-valid concepts. A snapshot of the FMEA-KG showing the sensor-asset-failure relationships is available at
\url{https://github.com/IBM/AssetOpsBench/blob/IndustryAssetEQA/IndustryAssetEQA/assets/KG.png}

\section{Appendix: Fact Extraction from Time Series Telemetry}
\label{sec:fact_extraction}

The fact-extraction stage converts raw timeseries data into deterministic, time-windowed episode facts suitable for ingestion into the episodic store. We provide a reproducible extractor. Key configurable parameters are the historical window length (\texttt{--window-hours}), the healthy-sampling horizon (\texttt{--horizon-hours}), and the maximum healthy episodes per machine (\texttt{--max-healthy-per-machine}). The extractor produces one JSON object per episode and writes them to a JSONL file (one JSON per line).

\subsection{Fact Extraction Procedure}
\begin{enumerate}[noitemsep, topsep=0pt, parsep=0pt, partopsep=0pt,
                leftmargin=1.5mm,    
                labelsep=1mm]      
  \item \textbf{Failure-centered episodes.} For each failure row (time $t$) the extractor builds the episode window $(t-\text{window\_hours},\, t]$ and aggregates telemetry, error, and maintenance records within that interval.
  \item \textbf{Healthy episodes.} For each machine the extractor subsamples telemetry timestamps (at most \texttt{max\_healthy\_per\_machine}). A candidate center $t$ is accepted as healthy if there is no failure in $[t,\ t+\text{horizon\_hours}]$, and we construct $(t-\text{window\_hours},\, t]$.
  \item \textbf{Feature engineering.} For each episode the extractor computes named scalar features:
    \begin{itemize}
      \item per-sensor aggregates and trend: \texttt{\{volt,rotate,pressure,vibration\}\_\\mean,std,min,max,trend};
      \item error aggregates: \texttt{error\_count\_last\_window}, \texttt{distinct\_error\_types\_last\_window};
      \item maintenance recency: \texttt{hours\_since\_last\_maint\_<component>} or \texttt{hours\_since\_last\_maint\_any};
      \item machine static: \texttt{machine\_age} and one-hot model flags \texttt{model\_<name>}.
    \end{itemize}
    Missing numeric values are emitted as JSON \texttt{null} (or a canonical sentinel such as \texttt{-1} for \texttt{hours\_since\_last\_maint} when no prior maintenance exists).
  \item \textbf{Provenance and identifiers.} Each fact includes a deterministic \texttt{fact\_id}, dataset and source filenames, telemetry time range, the failure row index (when applicable), and a \texttt{row\_index} for traceability back to raw records.
  \item \textbf{Output schema.} Facts are written to an output JSONL file. Each fact follows a compact schema (example below).
\end{enumerate}

\paragraph{CLI (example).}
The extractor was executed in our experiments with the following command (parameters shown):\\\\\\

\begin{samepage}
\begin{lstlisting}[style=cli,caption={Extractor CLI (example)},label={lst:cli_example}]
python -m src.utils.ts_fact_extractor \
  --telemetry ../msft_azure_pdm/PdM_telemetry.csv \
  --failures  ../msft_azure_pdm/PdM_failures.csv \
  --errors    ../msft_azure_pdm/PdM_errors.csv \
  --maint     ../msft_azure_pdm/PdM_maint.csv \
  --machines  ../msft_azure_pdm/PdM_machines.csv \
  --out       ../pdm_facts.jsonl \
  --window-hours 24 \
  --horizon-hours 24 \
  --max-healthy-per-machine 50
\end{lstlisting}
\end{samepage}

\subsection{Episodic Representation}
The extraction procedure yields a unified episodic representation. Across all datasets, each maintenance-relevant episode is encoded as a structured fact with a common schema.
\begin{itemize}[noitemsep, topsep=0pt, parsep=0pt, partopsep=0pt]
    \item \textbf{Asset identifiers}: asset instance, component (when applicable), and dataset source.
    \item \textbf{Episode type}: designation of failure-centered or healthy operation windows.
    \item \textbf{Temporal scope}: start and end timestamps defining the episode window, and failure time when applicable.
    \item \textbf{Engineered features}: summary statistics and trends computed from raw telemetry, along with error and maintenance recency indicators.
    \item \textbf{Observed label}: diagnostic state or failure mode associated with the episode.
    \item \textbf{Domain-aligned contexts}: asset, failure, and sensor profiles queried from FMEA Knowledge graph.
    \item \textbf{Provenance}: references to source files, record indices, and window-level statistics supporting verification and audit.
\end{itemize}

This explicit episodic representation serves as the single source of truth for question generation, answer verification, and evaluation, enabling temporal grounding, cross-dataset consistency, and reproducible, auditable embodied QA.

 Example single output fact (without KG enrichment), and output fact with KG enrichment produced by the extractor in our experiments are as follows:\\\\\\\\
 
\begin{lstlisting}[style=cli,language=bash,caption={Representative extracted fact  (without KG enrichment)},label={lst:fact_example}]

{"fact_id":"pdm_m56_comp3_2015-01-02T03",
 "dataset":"pdm","source_file":
 "PdM_telemetry.csv",
 "asset_id":"machine_56",
 "machineID":56,"episode_type":"failure_window",
 "failure_component":"comp3",
 "failure_time":"2015-01-02 03:00:00",
 "start_time":"2015-01-01 03:00:00","end_time":"2015-01-02 03:00:00",
 "label":"comp3",
 "features":[
   {"name":"volt_mean","value":169.0608},
   {"name":"volt_std","value":17.8556},
   {"name":"volt_min","value":139.2351},
   {"name":"volt_max","value":209.8819},
   {"name":"volt_trend","value":0.2177},
   ...,
   {"name":"hours_since_last_maint_comp3",
   "value":477.0},
   {"name":"machine_age","value":10.0},
   {"name":"model_model1","value":1.0}
 ],
 "provenance":{
   "telemetry_source_file":"PdM_telemetry.csv",
   "telemetry_time_range":["2015-01-01 03:00:00","2015-01-02 03:00:00"],
   "failure_source_file":"PdM_failures.csv",
   "failure_index":0,
   "errors_source_file":"PdM_errors.csv",
   "maint_source_file":"PdM_maint.csv",
   "machines_source_file":"PdM_machines.csv"
 },
 "row_index":0}
\end{lstlisting}

\begin{lstlisting}[style=cli, language=bash,caption={Example episodic fact that contains engineered features, provenance, and KG / FMEA enrichment (``asset\_profile'', ``failure\_profile'', ``sensor\_profiles'').}]
{"fact_id":"pdm_m73_comp4_2015-02-16T06",
"dataset":"pdm",
 "source_file":"PdM_telemetry.csv",
 "asset_id":"machine_73","machineID":73,
 "episode_type":"failure_window",
 "failure_component":"comp4",
 "failure_time":"2015-02-16 06:00:00",
 "start_time":"2015-02-15 06:00:00","end_time":"2015-02-16 06:00:00",
 "label":"comp4",
 "features":[
  {"name":"volt_mean","value":169.9217},
  {"name":"volt_std","value":13.1173},
  {"name":"volt_min","value":147.0643},
  {"name":"volt_max","value":193.2928},
  {"name":"volt_trend","value":0.2907},
  {"name":"rotate_mean","value":446.0634},
  {"name":"rotate_std","value":65.0448},
  {"name":"rotate_min","value":305.7116},
  {"name":"rotate_max","value":569.4687},
  {"name":"rotate_trend","value":-1.3637},
  {"name":"pressure_mean","value":99.8202},
  {"name":"pressure_std","value":10.2935},
  {"name":"vibration_mean","value":49.6625},
  {"name":"vibration_std","value":6.3090},
  {"name":"error_count_last_window",
  "value":1.0},
  {"name":"distinct_error_types_last_window",
  "value":1.0},
  {"name":"hours_since_last_maint_comp4",
  "value":5520.0},
  {"name":"machine_age","value":20.0},
  {"name":"model_model2","value":1.0}
 ],
 "asset_profile":{"asset_name":"model2"},
 "failure_profile":{
   "failure_label":"comp4","display_name":"comp4",
   "asset_name":"model2",
   "iso_metadata":{
     "failure_mode":"comp4",
     "name":"Rotor / bearing vibration fault",
     "description":"Mechanical wear, misalignment, or unbalance in the rotor, bearings, or coupling...",
     "equipment_category":"rotating_equipment",
     "associated_sensors":["vibration","rotate"],
     "typical_indicators":{
       "vibration_mean":"significantly above healthy baseline",
       "vibration_max":"high peaks"
     },
     "recommended_actions":[
       "perform vibration analysis and balancing",
       "inspect bearings, lubrication, and alignment",
       "check coupling condition and soft-foot or foundation issues"
     ],
     "severity":"very_high"
   }
 },
 "sensor_profiles":[
   {"sensor_name":"volt","description":"Sensors used to monitor the voltage..."},
   {"sensor_name":"rotate",
   "description":"Specialized sensors used to analyze vibrations..."},
   {"sensor_name":"pressure",
   "description":"Monitors the pressure..."},
   {"sensor_name":"vibration",
   "description":"Vibration sensors monitor machinery..."}
 ],
 "provenance":{
   "telemetry_source_file":"PdM_telemetry.csv",
   "telemetry_time_range":["2015-02-15 06:00:00","2015-02-16 06:00:00"],
   "failure_source_file":"PdM_failures.csv",
   "failure_index":120,
   "maint_events_in_window":1,
   "errors_in_window":1,
   "telemetry_points_in_window":24
 },
 "row_index":120}
\end{lstlisting}
\section{Appendix: Episodic Store (Ingest and Query layer)}
\label{sec:episodic_store}

The episodic store persists extracted episode facts and provides an API for deterministic retrieval, numeric feature queries, and verification used by the prompt builder and verifier. We implement a compact SQLite-backed store that stores the full JSON fact and an exploded features table for numeric search and ML export.

\subsection{Ingestion pipeline (example).}
After producing \texttt{../pdm\_facts.jsonl}, the store was created and populated using the following programmatic call (example shown as a one-liner for reproducibility):

\begin{lstlisting}[style=cli,language=bash,caption={Programmatic ingest into episodic store (example)},label={lst:episodic_ingest}]
python -c "from src.utils.episodic_store import EpisodicStore;
s = EpisodicStore('./pdm_episodic_store.db');
print('Ingested', s.ingest_jsonl('../pdm_facts.jsonl'));
print('Assets:', s.list_assets()); s.close()"
\end{lstlisting}

\noindent Example output from our run:
\begin{lstlisting}[style=cli, language=bash, caption={Episodic store ingest output (example)},label={lst:ingest_output}]
Ingested 5716
Assets: ['machine_1','machine_3',.,'machine_7']
\end{lstlisting}

\subsection{Schema and APIs.}
Each ingested episode is stored in a \texttt{facts} table and its named features are exploded into a \texttt{features} table. Key stored fields and API primitives:
\begin{itemize}[noitemsep, topsep=0pt, parsep=0pt, partopsep=0pt,
                leftmargin=1.5mm,    
                labelsep=1mm] 
  \item \textbf{Facts:} \texttt{fact\_id} (PK), \texttt{dataset}, \texttt{source\_file}, \texttt{asset\_id}, \texttt{row\_index}, \texttt{label}, \texttt{start\_time}, \texttt{end\_time}, \texttt{fact\_json} (full serialized fact), \texttt{ingest\_time}.
  \item \textbf{Features:} \texttt{fact\_id}, \texttt{feature\_name}, \texttt{feature\_value} (numeric if convertible), \texttt{feature\_text} (fallback for non-numeric entries).
  \item \textbf{APIs used by the pipeline:}
    \begin{itemize}
      \item \texttt{get\_fact(fact\_id)}: return the full fact JSON (used by prompt builder and verifier).
      \item \texttt{get\_features(fact\_id)}: return a name\(\rightarrow\)value/text mapping for prompt construction and ML export.
      \item \texttt{query\_by\_asset(asset\_id, limit)} and \texttt{query\_by\_label(label, limit)}: deterministic retrieval used in evaluation sampling and prompt contexts.
      \item \texttt{search\_by\_feature\_threshold \\ (feature\_name, operator (\(\bowtie\in\{<,\le,=,\ge,>\}\), value)}: numeric threshold search (e.g., \texttt{vibration\_max > 60}) for evidence selection.
      \item \texttt{query\_by\_time\_range(asset\_id, start, end)}: retrieve episodes overlapping a given time window (used by temporal QA and provenance checks).
      \item \texttt{export\_features\_csv(out\_csv)}: helper to materialize ML-ready CSVs.
    \end{itemize}
\end{itemize}

\subsection{Operational notes and reproducibility.}
\begin{itemize}
  \item Ingestion is idempotent (upsert semantics) and records an \texttt{ingest\_time} for traceability. Facts with the same \texttt{fact\_id} are replaced if \texttt{overwrite=True} is passed.
  \item Numeric conversions are attempted for feature values; non-numeric text is preserved in \texttt{feature\_text} so provenance and human inspection remain possible.
  \item The lightweight SQLite store is sufficient for experimental evaluation and reproducibility; it supports deterministic retrievals used in prompt construction and in automatic verification. For high-throughput production deployments a scalable store (e.g., a managed DB or vector store with feature indexing) would be recommended.
\end{itemize}

\subsection{Traceability example.}
The prompt builder and verifier rely on entries such as \texttt{telemetry\_time\_range} and \texttt{failure\_index} to show exactly which records support a claim. For example, given \texttt{fact\_id = "pdm\_m56\_comp3\_2015-01-02T03"}, the QA builder can:
\begin{enumerate}
  \item call \texttt{get\_fact(fact\_id)} to obtain the episode-level fact and its provenance;
  \item expose only the minimal fields required by a question (e.g., features and provenance) in the constructed prompt;
  \item after the LLM produces a response, the verifier can confirm that all cited \texttt{fact\_id}s, sensor windows, and referenced features exist in the store.
\end{enumerate}

\section{Appendix: Risk Modeling and Causal Simulator}
\label{sec:causal_sim}
The risk modeling and causal simulator provides a local, parametric counterfactual estimate of intervention effects; it is not a structural causal model and does not guarantee identifiability. The `do' is implemented by feature replacement (it ignores latent confounders and assumes the learned $P(y\mid \bm{x})$ remains valid under the intervention). Unknown intervention features are silently ignored, and the approach is parametric and model-dependent.

A simple confidence heuristic is used by the estimator:
\[
c \;=\; 0.5 + 0.5\min\!\Bigl(1,\; \frac{\lvert r_{\text{before}} - 0.5\rvert}{0.5}\Bigr),
\]
which increases when the baseline risk is more extreme (i.e., moves away from $0.5$).

\section{Appendix: Example QA Instances}
\label{app:examples}

This section provides representative JSON examples of generated QA instances (diagnostic, descriptive, temporal, counterfactual, action-oriented). Each QA shows how FMEA-KG fields (when available in the fact) are surfaced in the question, answer, reasoning, and provenance.

\subsubsection{Diagnostic (with KG excerpt)}
\begin{lstlisting}[style=cli, language=bash,caption={Diagnostic QA example}]
{"qa_id":"pdm_diag_pdm_m56_comp3_2015-01-02T03",
 "fact_id":"pdm_m56_comp3_2015-01-02T03",
 "task_type":"diagnostic",
 "question":"Why is this diagnostic episode for asset machine_56 labeled 'comp3' (comp3) over the time window 2015-01-01 03:00:00 to 2015-01-02 03:00:00?",
 "direct_answer":"This episode is labeled 'comp3' (comp3) because the observed features match the typical indicators associated with this failure mode.",
 "reasoning_answer":"This episode is labeled 'comp3' (comp3) because, key observed features: volt_mean=169.061, volt_std=17.856, volt_min=139.235. These features deviate from normal thresholds and sensor patterns and belong to the failure modes of a compressor",
 "provenance":{
   "fact_id":"pdm_m56_comp3_2015-01-02T03",
   "features":["volt_mean","volt_std",
   "volt_min"],
   "file":"PdM_telemetry.csv","row":0,
   "asset_profile":{"asset_name":"model1",
   "equipment_category":"rotating_equipment"},
   "failure_profile_id":156,
   "telemetry_points_in_window":22},
 "label":"comp3","asset_id":"machine_56",
 "asset_profile_brief":{"asset_name":"model1",
 "equipment_category":"rotating_equipment"},
 "failure_profile_brief":{
   "failure_mode":"Rotor / bearing vibration fault","display_name":"comp3",
   "severity":"very_high",
   "associated_sensors":["vibration"],
   "recommended_actions":["perform vibration analysis and balancing", "..."]}}
\end{lstlisting}

\subsubsection{Descriptive (sensor description surfacing)}
\begin{lstlisting}[style=cli, language=bash,caption={Descriptive QA example}]
{"qa_id":"pdm_desc_pdm_m56_comp3_2015-01-02T03",
 "fact_id":"pdm_m56_comp3_2015-01-02T03",
 "task_type":"descriptive",
 "question":"During the time window 2015-01-01 03:00:00 to 2015-01-02 03:00:00 for asset machine_56, what was the average vibration level (Vibration sensors monitor machinery for abnormal vibrations, which can indicate issues like misalignment, imbalance, or wear.)?",
 "direct_answer":"The average vibration level was approximately 39.26.",
 "reasoning_answer":"In this episode for asset machine_56, the feature vibration_mean is 39.26, computed over 22 telemetry points in the window.",
 "provenance":{"fact_id"
 :"pdm_m56_comp3_2015-01-02T03",
 "features":["vibration_mean"],
 "file":"PdM_telemetry.csv","row":0,
 "telemetry_points_in_window":22,
 "sensor_description":"Vibration sensors monitor machinery for abnormal vibrations..."},
 "label":"39.2568","asset_id":"machine_56"}
\end{lstlisting}

\subsubsection{Temporal / Counting}
\begin{lstlisting}[style=cli, language=bash,caption={Temporal / counting QA example}]
{"qa_id":"pdm_temp_pdm_m56_comp3_2015-01-02T03",
 "fact_id":"pdm_m56_comp3_2015-01-02T03",
 "task_type":"temporal_count",
 "question":"Between 2015-01-01 03:00:00 and 2015-01-02 03:00:00 for asset machine_56, how many distinct error types occurred?",
 "direct_answer":"There were 2 distinct error types in this window.",
 "reasoning_answer":"In this episode the feature distinct_error_types_last_window is 2, and the total error count is 2.",
 "provenance":{"fact_id":
 "pdm_m56_comp3_2015-01-02T03",
 "features":["error_count_last_window",
 "distinct_error_types_last_window"],
 "file":"PdM_errors.csv","row":0},
 "label":"0","asset_id":"machine_56"}
\end{lstlisting}

\subsubsection{Counterfactual (with KG fields in provenance)}
\begin{lstlisting}[style=cli, language=bash,caption={Counterfactual QA example}]
{"qa_id":"pdm_ts_cf_pdm_m56_comp3_2015-01-02T03",
 "fact_id":"pdm_m56_comp3_2015-01-02T03",
 "task_type":"counterfactual",
 "question":"For asset machine_56 (failure mode: comp3) in the window 2015-01-01 03:00:00 to 2015-01-02 03:00:00, if maintenance targeting comp3 had been performed immediately before the window (resetting hours_since_last_maint_comp3 from 477.0 to 0), how would the risk of failure change?",
 "direct_answer":"The risk of failure would decrease.",
 "reasoning_answer":"Baseline P(failure) $\approx$ 1.000; post-intervention P(failure) $\approx$ 0.000 (Δ = -1.000).",
 "provenance":{
   "fact_id":"pdm_m56_comp3_2015-01-02T03",
   "features":["hours_since_last_maint_comp3"],
   "file":"PdM_telemetry.csv","row":0,
   "telemetry_points_in_window":22,
   "asset_profile_brief":{"unit_subunit":
   
   ["rotor","bearings","coupling"],
   "asset_name":"model1"},
   "failure_profile_brief":{"failure_mode"
   :"comp3",
   "recommended_actions":["perform vibration analysis and balancing", "..."],"severity":"very_high"},
   "sensor_profiles_brief":
   [{"sensor_name":"vibration",
   "description":"..."}]
 },
 "counterfactual":{
   "intervention":"do
   (hours_since_last_maint_comp3 = 0.0)","risk_before":1.0,
   "risk_after":9.256e-06,"delta_risk":-0.99999,
   "direction":"decrease",
   "probs_before":{...},"probs_after":{...}
 },
 "confidence":1.0,"label":"comp3",
 "asset_id":"machine_56"}
\end{lstlisting}

\subsubsection{Action recommendation (with KG fields in provenance)}
\begin{lstlisting}[style=cli, language=bash,caption={Action-recommendation QA example}]
{"qa_id": "pdm_ts_action_"
"pdm_m56_comp3_2015-01-02T03",
"fact_id": "pdm_m56_comp3_2015-01-02T03",
"task_type": "action_recommendation",
"question": "For rotating_equipment machine_56 in the time window 2015-01-01 03:00:00 to 2015-01-02 03:00:00, should a maintenance work order be opened now, or is it acceptable to continue monitoring?",
"direct_answer": "A maintenance work order should be opened now.",
"reasoning_answer": "The learned risk model estimates probability of any failure $\approx$ 1.00 (threshold=0.50). Failure severity: high. Recommended diagnostics/actions: check discharge and downstream valves for incorrect positions or sticking; inspect filters, strainers, and lines for blockage; review process conditions and control setpoints. Most informative sensors: pressure. Equipment class/type: electric_motor_driven_rotating_machine.",
"provenance": {
"fact_id": "pdm_m56_comp3_2015-01-02T03",
"features": ["volt_mean", "volt_std", "volt_min", "volt_max", "volt_trend",
"rotate_mean", "rotate_std", "rotate_min", "rotate_max", "rotate_trend",
"pressure_mean", "pressure_std", "pressure_min", "pressure_max", "pressure_trend",
"vibration_mean", "vibration_std", "vibration_min", "vibration_max", "vibration_trend",
"error_count_last_window", "distinct_error_types_last_window",
"hours_since_last_maint_comp3", "machine_age", "model_model1"],
"file": "PdM_telemetry.csv",
"row": 0,
"telemetry_points_in_window": 22,
"errors_in_window": 0,
"failure_profile_id": "comp3",
"asset_profile_brief": {
"equipment_category": "rotating_equipment",
"equipment_class_type": "electric_motor_driven_rotating_machine",
"unit_subunit": ["process_path", "discharge_line", "valves"],
"asset_name": "model1"
},
"failure_profile_brief": {
"failure_mode": "comp3",
"display_name": "comp3",
"short_description": "Restriction, valve malfunction, or process-side blockage causing sustained high discharge pressure while rotational speed and vibration remain near normal.",
"associated_sensors": ["pressure"],
"typical_indicators": {
"pressure_mean": "significantly above healthy baseline",
"pressure_trend": "often positive or sustained high",
"rotate_mean": "near nominal",
"vibration_mean": "near nominal or slightly increased"
},
"recommended_actions": ["check discharge and downstream valves for incorrect positions or sticking",
"inspect filters, strainers, and lines for blockage",
"review process conditions and control setpoints"],
"severity": "high"
}
},
"label": "open_work_order",
"asset_id": "machine_56",
"confidence_estimator": 1.0,
"risk": 1.0,
"probs_before": {
"comp1": 5.666517784187859e-26,
"comp2": 1.6855443558067901e-53,
"comp3": 1.0,
"comp4": 4.746799049527151e-27,
"healthy": 1.1467891488958534e-43
},
"asset_profile_brief": {
"equipment_category": "rotating_equipment",
"equipment_class_type": "electric_motor_driven_rotating_machine",
"unit_subunit": ["process_path", "discharge_line", "valves"],
"asset_name": "model1"
},
"failure_profile_brief": {
"failure_mode": "comp3",
"display_name": "comp3",
"short_description": "Restriction, valve malfunction, or process-side blockage causing sustained high discharge pressure while rotational speed and vibration remain near normal.",
"associated_sensors": ["pressure"],
"typical_indicators": {
"pressure_mean": "significantly above healthy baseline",
"pressure_trend": "often positive or sustained high",
"rotate_mean": "near nominal",
"vibration_mean": "near nominal or slightly increased"
},
"recommended_actions": ["check discharge and downstream valves for incorrect positions or sticking",
"inspect filters, strainers, and lines for blockage",
"review process conditions and control setpoints"],
"severity": "high"
}}
\end{lstlisting}

\section{Appendix: Prompt Builder and Prompt Templates.}
\label{sec:prompt_builder}

Below is a compact version of the template used in our experiments (user prompt shown; the system prompt contains the role and the output contract):

\begin{lstlisting}[style=cli, language=bash,caption={Example prompt generated}]
TASK TYPE:Diagnostic

TASK DESCRIPTION:
You must answer diagnostic questions explaining why the episode has the given label.
Use only the evidence; do not invent features or events.

EVIDENCE:
fact_id: pdm_m56_comp3_2015-01-02T03
asset_id: machine_56
window_start: 2015-01-01 03:00:00
window_end:   2015-01-02 03:00:00
diagnostic_features:
  - vibration_mean: 39.26
  - pressure_mean: 124.67
  - hours_since_last_maint_comp3: 477.0
asset_profile:
  asset_name: model1
failure_profile:
  name: High discharge pressure / process restriction
  associated_sensors: [pressure]

QUESTION:
Why is this episode labeled as failure for asset 'compressor 3'?

RESPONSE FORMAT:
Return ONLY a single JSON object with keys:
  direct_answer, reasoning_answer, provenance, confidence
\end{lstlisting}

\begin{lstlisting}[style=cli, language=bash, caption={Prompt Builder CLI (example)}]
python -m src.utils.prompt_builder_static \
  --db ../pdm_episodic_store.db \
  --qa ../pdm_qa_diag.jsonl \
  --qa-id pdm_diag_pdm_m56_comp3_2015-01-02T03
\end{lstlisting}

\section{Appendix: Datasets Used}
\label{sec:datasets}
\paragraph{Microsoft Azure Predictive Maintenance Dataset (PdM).}
The \emph{Microsoft Azure Predictive Maintenance dataset}\cite{azure_pdm_kaggle} is a multi-table corpus designed for developing and evaluating ML models for predictive maintenance. It contains:
\begin{itemize}[noitemsep,leftmargin=*]
  \item Hourly telemetry (voltage, rotation, pressure, vibration) for 100 machines: \texttt{PdM\_telemetry.csv}.
  \item Failure log (component replacements after breakdowns): \texttt{PdM\_failures.csv}.
  \item Non-fatal error events (do not cause shutdown): \texttt{PdM\_errors.csv}.
  \item Maintenance records (scheduled pro-active and unscheduled reactive interventions): \texttt{PdM\_maint.csv}.
  \item Machine metadata (model, age, etc.): \texttt{PdM\_machines.csv}.
\end{itemize}

Times in all tables are rounded to the nearest hour to align with the telemetry sampling; failures are a subset of maintenance events (component replacements following a breakdown). The dataset therefore supports episode construction that pairs fixed historical telemetry windows with downstream outcomes (failure vs.\ healthy) and contextual covariates (error counts, hours-since-last-maintenance, machine model/age), making it well suited for supervised risk modelling, time series feature engineering, and counterfactual or what-if analyses in predictive-maintenance research.

\paragraph{Turbofan Engine Degradation Dataset (C-MAPSS).}
We use the \emph{NASA C-MAPSS (Commercial Modular Aero-Propulsion System Simulation) turbofan engine dataset} \cite{saxena2008damage, nasa_cmapss_opendata}, a widely adopted benchmark for prognostics and health management. The dataset consists of multivariate run-to-failure time series collected from a fleet of simulated turbofan engines, where each trajectory corresponds to a distinct engine with unknown initial wear and manufacturing variability. Engines operate under varying operational settings and are monitored via multiple sensor channels corrupted by realistic noise. Faults emerge during operation and progressively grow until system failure in the training sets, while test trajectories terminate prior to failure with ground-truth Remaining Useful Life (RUL) values provided. 

We use the training and test splits of all four standard subsets: FD001 (100/100 trajectories, single operating condition, single fault mode: high-pressure compressor degradation), FD002 (260/259 trajectories, six operating conditions, single fault mode), FD003 (100/100 trajectories, single operating condition, two fault modes: high-pressure compressor and fan degradation), and FD004 (248/249 trajectories, six operating conditions, two fault modes). Each time step represents one operational cycle and includes three operational settings and 21 sensor measurements, yielding a total of 26 variables per cycle. For IndustryAssetEQA, raw sequences are segmented into fixed-length episodic windows aligned with degradation progression, enabling descriptive, temporal, diagnostic, counterfactual, and action-oriented question answering over realistic degradation dynamics.

\paragraph{Cyber-Physical Production Systems (Genesis).}
\emph{Genesis cyber-physical production system} \cite{von2018anomaly} captures sensor-level behavior of an industrial demonstrator that autonomously sorts two materials (wood and metal) into designated storage locations. The system consists of four modular stations—storage magazine, sensor module, wood storage, and metal storage—whose physical positions can be permuted, with the PLC control logic automatically adapting to the configuration. Material transport is carried out by a linear drive equipped with a pneumatic gripper, executing a cyclic procedure that includes homing, material ejection, gripping, material-type sensing, transport to the appropriate storage box, and release. The dataset comprises two complementary subsets. 

The first contains fully labeled multivariate time series data (\texttt{Genesis\_StateMachineLabels.csv}, \texttt{Genesis\_AnomalyLabels.csv}) with 16{,}220 observations sampled every \(50\,\mathrm{ms}\) via an OPC DA server. These files share identical sensor readings but differ in labeling: one annotates the internal PLC state machine (idle, homing, transport, gripping, sensing, and release states), while the other provides manually verified anomaly labels corresponding to linear drive malfunctions (jammed/tilted motion and subsequent correction). The second subset consists of unlabeled runs (\texttt{Genesis\_normal.csv}, \texttt{Genesis\_lineardrive.csv}, \texttt{Genesis\_pressure.csv}) capturing nominal operation and gradually induced faults in the linear drive and pneumatic pressure. For IndustryAssetEQA, we derive short episodic windows from these time series to evaluate descriptive, temporal, and diagnostic question answering under realistic cyber-physical noise and partial supervision, complementing the more failure-centric PdM and C-MAPSS datasets.

\paragraph{Hydraulic Systems Dataset.}
We additionally evaluate IndustryAssetEQA on the \emph{Condition Monitoring of Hydraulic Systems} dataset \cite{hydraulic_zenodo_1323611, helwig2015condition, helwig2015d8, condition_monitoring_of_hydraulic_systems_447}, which captures multivariate sensor measurements from a laboratory hydraulic test rig under controlled degradation. The rig consists of a primary working circuit and a secondary cooling–filtration circuit connected via a shared oil tank, and repeatedly executes constant-load operating cycles of 60 seconds. Each cycle is annotated with condition states of four hydraulic components—cooler, valve, pump, and accumulator—across multiple severity levels.

The dataset contains 2,205 operating cycles with no missing values. Each cycle is represented as a matrix of raw time series measurements collected at heterogeneous sampling rates, yielding 43,680 attributes per instance. Sensors include pressures (PS1–PS6, 100 Hz), motor power (EPS1, 100 Hz), volume flow (FS1–FS2, 10 Hz), temperatures (TS1–TS4, 1 Hz), vibration (VS1, 1 Hz), and virtual efficiency indicators (CE, CP, SE, 1 Hz). Component condition labels are provided cycle-wise in a separate profile file and include both categorical degradation states (e.g., leakage levels) and continuous-valued health indicators (e.g., accumulator pressure).

Following prior work, we treat each operating cycle as a maintenance-relevant episode and aggregate sensor traces into fixed-length episodic representations suitable for descriptive, diagnostic, temporal, and counterfactual question answering. This dataset is widely used for benchmarking condition monitoring and fault diagnosis methods and provides realistic, multi-rate telemetry for evaluating evidence-grounded industrial QA.





\section{Appendix: Evaluation}
\label{sec:evaluation}

\subsection{Baselines}
\label{sec:baselines}
We compare IndustryAssetEQA against several baselines which we term as system configurations that reflect common design choices in industrial QA systems. While several baseline configurations correspond to partial versions of IndustryAssetEQA, we treat baselines as externally reasonable system designs for comparison, whereas ablations explicitly remove components from the full IndustryAssetEQA system to isolate causal contributions.

\paragraph{LLM-only QA.}
The LLM is prompted with the question and a natural-language summary of the episode, without structured evidence, provenance constraints, and simulator outputs.

\paragraph{LLM with Episodic Evidence.}
The model receives a textual rendering of episode features and time windows, but is not required to produce structured outputs or explicit provenance.

\paragraph{LLM with Episodic Evidence + KG Grounding.}
In addition to episodic facts, the model is provided with structured domain knowledge derived from FMEA-KG. This includes failure mode metadata, equipment categories and classes, unit and subunit types, sensor–failure associations, and recommended maintenance actions. Outputs remain unconstrained free text, and no verification is applied. This variant isolates the effect of symbolic domain knowledge grounding independent of structural and verifier-based constraints.

\paragraph{Provenance-Enforced QA (no simulator).}
The model is required to return structured JSON outputs containing an explicit \texttt{provenance} field that cites episode identifiers and feature names. A deterministic verifier validates these references against the Episodic Store and rejects any output with missing, malformed, or unsupported evidence. This configuration enforces evidence correctness post-generation but does not provide access to the risk simulator for counterfactual reasoning.

\paragraph{Full IndustryAssetEQA.}
The complete system integrates episodic grounding, KG-based domain knowledge, provenance-enforced structured outputs, and simulator-grounded counterfactual reasoning. This configuration reflects the intended deployable system.

\subsection{Evaluation Metrics}
\label{sec:metrics}
We define the following evaluation metrics, all of which are computed automatically using deterministic verifiers over structured model outputs.

\paragraph{Structural Validity (\texttt{Struct.OK}).}
This metric checks whether the model output conforms to the required JSON schema, including the presence and type correctness of mandatory fields such as \texttt{direct\_answer}, \texttt{reasoning\_answer}, \texttt{provenance}, and \texttt{confidence}. Outputs that fail structural validation are considered unusable in deployed systems and are counted as failures.

\paragraph{Provenance Accuracy (\texttt{Prov.OK}).}
Provenance accuracy verifies that the response cites a valid episodic fact identifier, references only features that actually exist in that episode, and matches the recorded source file and row index when provided. This check is implemented by validating model outputs against the Episodic Store and penalizes hallucinated or unsupported evidence.

\texttt{Struct.OK} and \texttt{Prov.OK}  are implemented by the \texttt{Verifier module} that validates model outputs against the episodic store.

\paragraph{Label Consistency (\texttt{Label Cons.}).}
For diagnostic QA tasks, \texttt{label\_consistent} measures agreement between the model’s predicted diagnostic label and the SME-validated reference label. To account for terminology variation, labels are normalized using the FMEA knowledge graph, allowing matches via synonyms or hierarchical relations. This metric evaluates final-answer correctness only and does not assess the correctness or faithfulness of the generated explanation. Label consistency is not applicable to non-diagnostic tasks.

\paragraph{Full Pass Rate (\texttt{Full\_pass}).}
The full pass rate measures the fraction of QA instances for which all applicable checks succeed simultaneously: structural validity, provenance accuracy, and label consistency. This metric serves as a conservative proxy for \emph{deployable answer reliability}, as any failure indicates an answer that cannot be safely consumed by downstream systems.

\paragraph{Temporal and Counting Accuracy.}
For temporal and counting QA tasks, we evaluate whether predicted timestamps, durations, or counts match the episode-derived ground truth. Depending on the task, this is computed as exact match or window-level overlap.

\paragraph{Counterfactual Direction Accuracy (\texttt{CF Acc.}).}
For counterfactual QA tasks, we evaluate whether the predicted direction of risk change (\emph{increase}, \emph{decrease}, or \emph{no change}) matches the direction produced by the parametric risk simulator, which we use as the surrogate intervention model. The simulator’s directional predictions were reviewed by domain experts on sampled cases for directional plausibility. We therefore measure agreement with the expert-reviewed surrogate model rather than claiming causal ground truth under real-world interventions. When numeric risk estimates (\texttt{risk\_before} and \texttt{risk\_after}) are present in the model output, the direction is computed directly from these values; textual direction labels are used only as a fallback. This metric evaluates action–model consistency under an explicit parametric intervention assumption rather than empirical causal validity.

\paragraph{Entailment Pass Rate (\texttt{Entail.Pass}).}
This metric measures whether generated explanations are supported by the provided evidence. Each reasoning answer is segmented into individual sentences, and a natural language inference (NLI) model, \texttt{FacebookAI
/roberta-large-mnli} \cite{liu2019roberta} is used to test whether each sentence is entailed by the episodic facts and FMEA knowledge supplied in the prompt. A sentence is counted as entailed if its entailment probability exceeds a fixed threshold. 
We fix the threshold at 0.80 following prior empirical evaluations. The entailment pass rate is computed as the fraction of reasoning sentences that satisfy this criterion, capturing the degree to which explanations avoid unsupported or contradictory claims.

\paragraph{Claim Precision (\texttt{Claim Prec.}).}
Claim precision evaluates the factual correctness of explanations at a symbolic level. Atomic claims are extracted from the reasoning text (e.g., assertions about feature states or failure–symptom relationships) and verified deterministically against the episodic store and the FMEA knowledge graph. Claim precision is defined as the fraction of extracted claims that can be validated. This metric penalizes explanations that introduce incorrect or unverifiable statements, even when the final answer is correct.


\subsection{Ablation Design}
\label{sec:ablation}
To understand which components materially affect model behavior, we perform controlled ablations:
\begin{itemize}
    \item \textbf{No episodic memory}: remove the episodic store and explicit episode identifiers, providing only free-form summaries of telemetry without fact-level indexing or verifiable provenance.

    \item \textbf{No FMEA knowledge graph}: remove KG-derived semantic context (failure-mode metadata, associated sensors, and recommended actions), leaving the model to rely solely on episodic telemetry and prompts.
    \item \textbf{No provenance enforcement}: remove the structured output requirement and verifier checks, allowing free-text explanations without machine-verifiable provenance.

    \item \textbf{No risk simulator}: require the model to answer counterfactual questions without simulator guidance.
\end{itemize}

All models are evaluated on the same QA instances to isolate the effect of architectural constraints rather than dataset differences.

\end{document}